\documentclass[lettersize,journal]{IEEEtran}
\usepackage{amsmath,amsfonts}
\usepackage{algorithmic}
\usepackage{algorithm}
\usepackage{array}
\usepackage[caption=false,font=normalsize,labelfont=sf,textfont=sf]{subfig}
\usepackage{textcomp}
\usepackage{stfloats}
\usepackage{url}
\usepackage{verbatim}
\usepackage{graphicx}
\usepackage{makecell}
\usepackage{multirow}
\usepackage{multicol}
\usepackage{booktabs}
\usepackage{amsmath}
\usepackage{threeparttable}
\usepackage[table]{xcolor}
\usepackage{hyperref}
\hypersetup{
    colorlinks=true,
    linkcolor=blue,
    urlcolor=blue,
    citecolor=blue}
\usepackage[T1]{fontenc}

\begin{document}

\title{Enhancing Underwater Images via Adaptive Semantic-aware Codebook Learning}

\author{Bosen Lin, 
Feng Gao, \emph{Member}, \emph{IEEE},
Yanwei Yu, \emph{Member}, \emph{IEEE},
Junyu Dong, \emph{Member}, \emph{IEEE}, \\
Qian Du, \emph{Fellow}, \emph{IEEE}
\thanks{This work was supported in part by the National Science and Technology Major Project of China under Grant 2022ZD0117202, in part by the Natural Science Foundation of Shandong Province under Grant ZR2024MF020. (\textit{Corresponding author: Feng Gao})

Bosen Lin, Feng Gao, Yanwei Yu, and Junyu Dong are with the School of Information Science and Engineering, Ocean University of China, Qingdao 266100, China. 

Qian Du is with the Department of Electrical and Computer Engineering, Mississippi State University, Starkville, MS 39762 USA.}}

\markboth{IEEE TRANSACTIONS ON GEOSCIENCE and REMOTE SENSING}{}

\maketitle

\begin{abstract}
Underwater Image Enhancement (UIE) is an ill-posed problem where natural clean references are not available, and the degradation levels vary significantly across semantic regions. Existing UIE methods treat images with a single global model and ignore the inconsistent degradation of different scene components. This oversight leads to significant color distortions and loss of fine details in heterogeneous underwater scenes, especially where degradation varies significantly across different image regions. Therefore, we propose SUCode (Semantic-aware Underwater Codebook Network), which achieves adaptive UIE from semantic-aware discrete codebook representation. Compared with one-shot codebook-based methods, SUCode exploits semantic-aware, pixel-level codebook representation tailored to heterogeneous underwater degradation. A three-stage training paradigm is employed to represent raw underwater image features to avoid pseudo ground-truth contamination. Gated Channel Attention Module (GCAM) and Frequency-Aware Feature Fusion (FAFF) jointly integrate channel and frequency cues for faithful color restoration and texture recovery. Extensive experiments on multiple benchmarks demonstrate that SUCode achieves state-of-the-art performance, outperforming recent UIE methods on both reference and no-reference metrics. The code will be made public available at \url{https://github.com/oucailab/SUCode}.
\end{abstract}


\section{Introduction}
Underwater vision plays a crucial role in extending remote sensing capabilities to marine and freshwater environments, which are otherwise difficult to observe using traditional aerial or satellite sensors \cite{wenSemiSupervisedDomainAdaptiveFramework2025a}\cite{liuHDANetEnhancingUnderwater2025}\cite{yuTaskFriendlyUnderwaterImage2024}. High-quality underwater imaging is critical for autonomous underwater robots inspection, deep-sea archaeology, and marine life monitoring. Unfortunately, wavelength-dependent absorption and scattering effects in water make it exceptionally difficult to capture visually reliable images in situ. To address this issue, various Underwater Image Enhancement (UIE) methods have been proposed to restore color fidelity and structural details \cite{zhangSynergisticMultiscaleDetail2024} \cite{pengAdaptiveDualdomainLearning2025} \cite{changWaterDiffusionLearningPriorinvolved2025}.

\begin{figure}[h]
  \centering
  \includegraphics[width=0.8\linewidth]{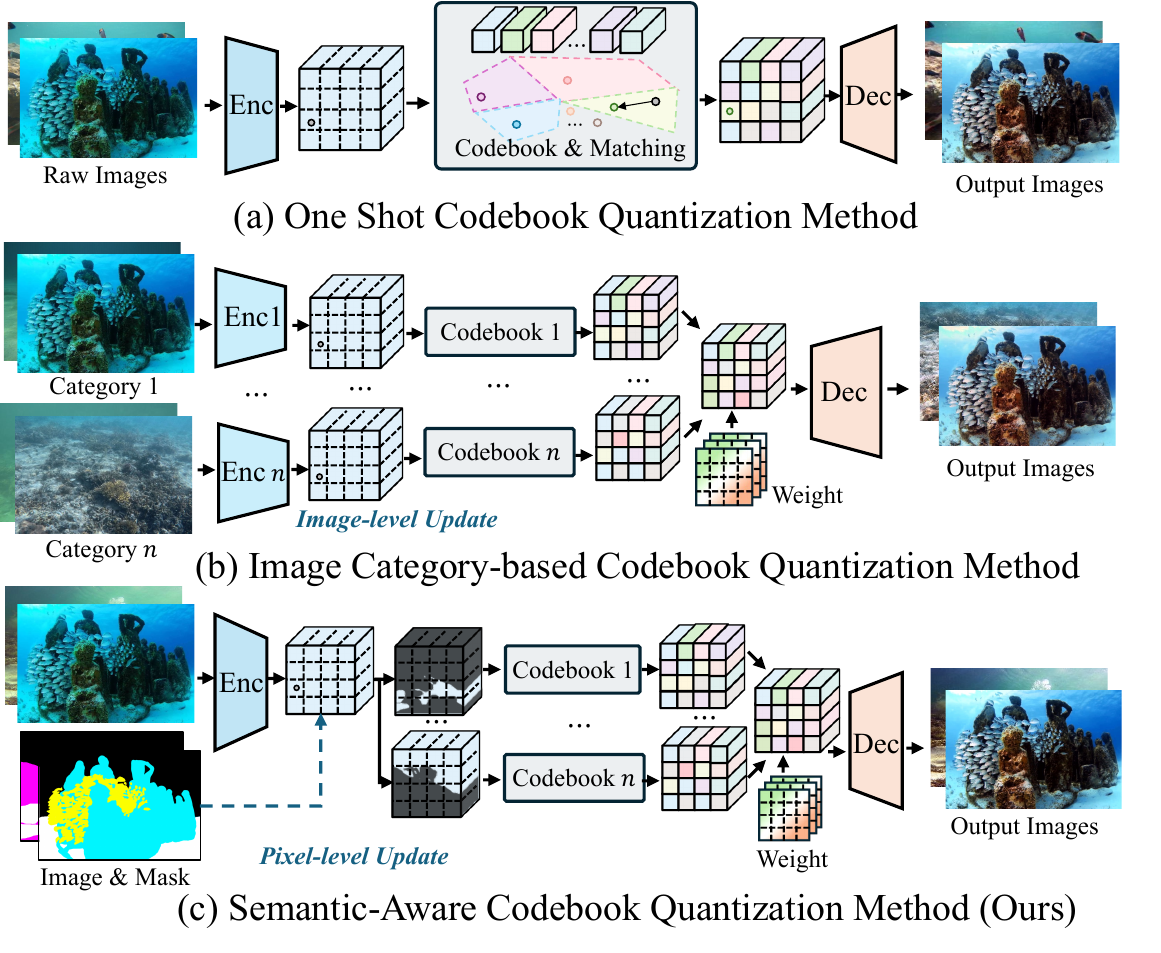}
  \caption{The comparison of the training and testing pipeline and enhance results between different codebook generation methods. The proposed SUCode's result is sharper and clearer, with more natural color. }
  \label{Figure:introduction}
\end{figure}

However, UIE faces two significant challenges: \textit{\textbf{(1) Ill-posed problem and absence of clean ground-truth reference}}: Unlike natural or remote sensing image enhancement tasks, where ground-truth images serve as ideal baselines, UIE lacks truly clean reference images. In natural image enhancement, degraded images can be directly compared with clean ground truth to guide learning and evaluation \cite{liuLearningImageAdaptiveCodebooks2023}. In contrast, underwater imaging suffers from unavoidable degradation including turbidity, light absorption, scattering, and fluctuating water currents. These factors lead to inconsistent color shifts, contrast loss, and attenuation of details in unpredictable ways \cite{pengUShapeTransformerUnderwater2023}. These effects vary across environments, making it impossible to capture a distortion-free underwater scene. As a result, ``ground truth" references for UIE are pseudo ground-truths, typically chosen by human annotators from multiple enhanced outputs. This lack of reliable references complicates both the training and objective evaluation of UIE methods.
\textit{\textbf{(2) Highly variable degradation patterns across depth, turbidity, and scene content:}} Degradation in underwater images is non-uniform and strongly influenced by water depth, turbidity levels, and scene composition. For instance, foreground objects closer to the camera usually receive better illumination, while deeper background regions suffer severe color distortion and contrast loss. Additionally, dynamic disturbances such as floating particles, aquatic life, and water currents affect how light interacts with the water, introduce additional motion blur, noise artifacts, and structural inconsistencies. These variations cause non-uniform degradation both within individual frames and across consecutive frames, presenting a challenge for creating a universal enhancement solution. Conventional approaches often apply a single global mapping to enhance the entire frame \cite{liUnderwaterImageEnhancement2020} \cite{liUnderwaterImageEnhancement2021} \cite{zhangSynergisticMultiscaleDetail2024}, which tends to over-enhanced foregrounds while leaving haze in the background. To address this, it is essential to design robust UIE methods that can adapt to the spatially varying degradation encountered in real underwater scenarios.

Discrete generative priors facilitate the learning of a more compact and expressive latent space by mapping continuous image features to a discrete vector space, often represented as a Codebook. This discrete representation effectively captures essential information (such as color, texture, etc.) while eliminating redundant or noisy elements in the image. Such methods have demonstrated impressive performance in tasks like image denoising, deblurring, and restoration \cite{wuRIDCPRevitalizingReal2023a} \cite{zhouGLARELowLight2024} \cite{fuIterativePredictorCriticCode2025}. These approaches learn a high-quality codebook through self-reconstruction, then retrieve the most appropriate code entries to guide the restoration process.  However, directly applying this paradigm to UIE is non-trivial. First, UIE is an ill-posed problem, which lead to the learned high-quality codebook containing unpredictable noise, which negatively impacts enhancement performance. Second, the complex and inconsistent degradation patterns in underwater scenes challenge one-shot vector quantization strategies, such as CodeUNet \cite{wangCodeUNetAutonomousUnderwater2024} and AdaCode \cite{liuLearningImageAdaptiveCodebooks2023}, make them insufficient to capture diverse distortions. Consequently, enhancement networks often fail to align with the correct priors, as illustrated in Fig. \ref{Figure:introduction}(a).

In order to introduce the advantages of codebook learning into the field of UIE and simultaneously address spatially varying degradation and the ill-posed ground-truth contamination, we propose \textbf{S}emantic-aware \textbf{U}nderwater \textbf{Code}book Network (\textbf{SUCode}). SUCode is designed to restore clear underwater images from semantically relevant discrete raw image representations. In our approach, a series of pixel-level discrete codebooks in the latent space guided by semantic masks are first learned. Each codebook corresponds to a distinct raw underwater category. Next, a learnable weight predictor then linearly aggregates these basis codebooks to form an image-adaptive discrete  generative codebook for any given underwater image. Finally, a dual-decoder architecture reconstructs clear images that respect both global semantics and local details,  with Gated Channel Attention Module (GCAM) and Frequency-Aware Feature Fusion (FAFF). The GCAM enhances codebook representations by adaptively re-weighting color channels, while FAFF bridges raw and enhanced features in the frequency domain, further ensuring semantic consistency and detail preservation. To our knowledge, the proposed SUCode is the first network to incorporate semantic information into codebook learning in  UIE tasks, with its network architecture depicted in Fig. \ref{Figure:introduction}(a).

In summary, our main contributions include:
\begin{itemize}

\item We propose a novel codebook learning approach to obtain semantic-dependent discrete representations of UIE tasks. Unlike one-shot methods, our approach propose pixel-level, category-specific codebooks for underwater scenes guided by semantic masks. 

\item By learning directly from raw underwater images, SUCode addresses the ill-posed nature of UIE and avoids pseudo ground-truth contamination. Domain conversion to clear images is performed via a GCAM-enhanced and FAFF-bridged dual‑decoder architecture.

\item Extensive experiments on four public benchmarks demonstrate that SUCode outperforms nine state-of-the-art UIE models on full-reference metrics and achieves competitive results on no-reference metrics, while generalizing well to real-world UIEs without retraining.

\end{itemize}

\section{Related Work}\label{Section:2}

\subsection{Underwater Image Enhancement}
Traditional UIE methods often rely on physical models of underwater imaging and use hand-crafted priors and parameterized models for enhancement. These priors include the Underwater Light Attenuation Prior (ULAP) \cite{songEnhancementUnderwaterImages2020}, Image Blurriness and Light Absorption Prior (IBLA) \cite{pengUnderwaterImageRestoration2017a}, and the Underwater Dark Channel Prior \cite{houNovelDarkChannel2020}, among others. However, due to the significant variation in underwater imaging conditions, these priors struggle to model the complexities of real-world scenarios.

In recent years, deep learning methods have greatly advanced the development of UIE. These methods leverage transformers, generative models, and diffusion models to tackle the specific degradation present in underwater environments. WaterNet \cite{liUnderwaterImageEnhancement2020} learns a fusion representation of different traditional UIE techniques. UColor \cite{liUnderwaterImageEnhancement2021} employs a medium transmission-guided multi-color space embedding network, which jointly captures illumination, chromatic, and perceptual priors. Peng et al. \cite{pengUShapeTransformerUnderwater2023} designed a UNet-shaped Transformer network for end-to-end underwater image enhancement. Wf-Diff \cite{zhaoWaveletbasedFourierInformation2024} enhances the frequency information of underwater images in the wavelet space using a diffusion model. AMSIN \cite{quanEnhancingUnderwaterImages2024a} treats UIE as an image decomposition problem and designs a deep invertible neural network for bidirectional reversible enhancement. SS-UIE \cite{pengAdaptiveDualdomainLearning2025} employs a Mamba-based network and frequency domain loss to emphasize the enhancement of high-frequency details in underwater images. SMDR-IS \cite{zhangSynergisticMultiscaleDetail2024} utilizes multi-scale information in an encoder-decoder architecture to enhance underwater image details. However, these methods typically treat the image as a whole, without leveraging the semantic regions of underwater images for targeted enhancement. This limitation can result in enhancements that are not optimal for human visual perception. 

\begin{figure*}[h]
  \centering
  \includegraphics[width=0.8\linewidth]{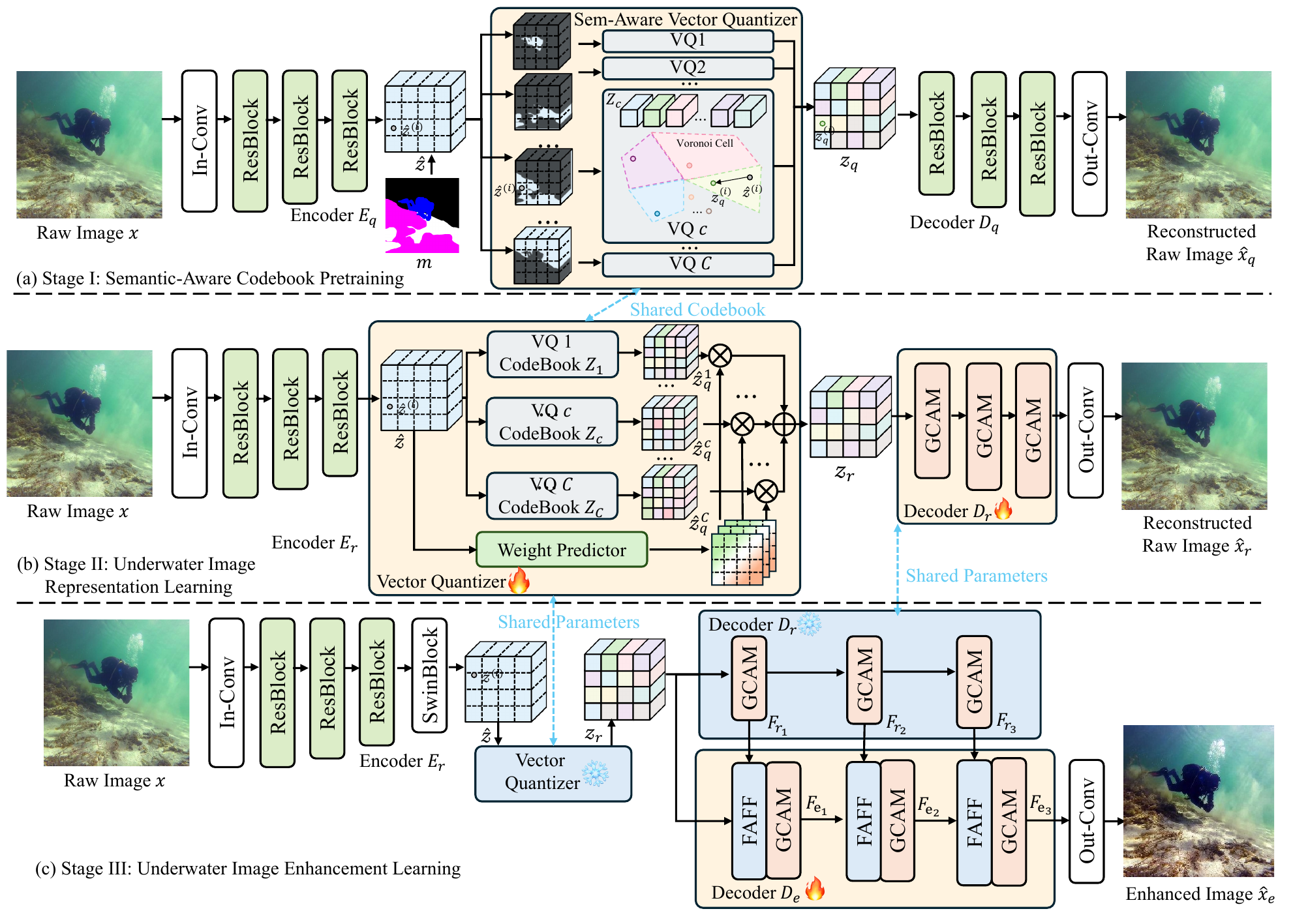}
  \caption{The overall structure of the proposed SUCode. In stage I, the semantic-aware category‑specific codebooks are updated with the mask $m$. Stage II is a partition and synthesis process of the codebook, achieved through the self-reconstruction of raw underwater images. In stage III, domain conversion from the raw underwater codebook to the enhanced image is achieved.  }
  \label{Figure:overallstructure}
\end{figure*}

\subsection{Codebook learning}
The concept of codebook learning was first introduced in VQ-VAE \cite{vandenoordNeuralDiscreteRepresentation2017}, a vector-quantized autoencoder framework that allows models to learn discrete feature representations in latent space, known as codebooks. VQGAN \cite{esserTamingTransformersHighResolution2021} further improves codebook learning by adding perceptual and adversarial supervision, which leads to a smaller codebook size and better reconstruction quality. Building on this, many low-level vision tasks have adopted discrete codebooks to enhance feature representation, including super-resolution, image restoration, and image dehazing. For instance, FeMaSR \cite{chenRealWorldBlindSuperResolution2022} applies discrete codebook learning to blind super-resolution. AdaCode \cite{liuLearningImageAdaptiveCodebooks2023} extends the robustness of class-agnostic codebooks by computing a weighted combination of multiple basis codebooks. RIDCP \cite{wuRIDCPRevitalizingReal2023a} introduces a controllable matching operation to dehaze low-quality images using high-quality codebook priors. IPC-Dehaze \cite{fuIterativePredictorCriticCode2025} gradually refines and retains the best high-quality codebook for image dehazing.

Despite the progress in codebook learning, few studies have explored its application in UIE. CodeUNet \cite{wangCodeUNetAutonomousUnderwater2024} follows the RIDCP framework and proposes prior codebooks specifically for water bodies to restore underwater images. VQUIE-Net \cite{dingVectorQuantizedUnderwater2025} leverages VQGAN to map underwater images into a discrete domain and employs an axial flow-guided latent transformer for image restoration. However, a key challenge remains: how to fully utilize the advantages of codebook learning to represent the complex foreground-background interactions in underwater scenes and address the problem of prior learning due to the ill-posed nature of ground-truth images. Therefore, it is necessary to improve the mapping from original image to enhanced one based on learning the representation of original underwater image.



\section{Methodology}\label{Section:3}
The motivation for SUCode is to introduce discrete representation into UIE tasks, where the key issues need to be addressed are the complex degradation patterns and the ill-posed references. Specifically, to capture the complex region-specific degradation typical of underwater imagery, the proposed SUCode introduces an additional semantic dimension to the standard discrete codebook representation \cite{esserTamingTransformersHighResolution2021}. To avoid learning unstable representations from ill-posed pseudo ground-truths, SUCode employs a progressive three-stage training strategy, which aims to learn quantized representations of raw underwater image content and treats the enhancement process as an adaptive feature modulation problem.

As shown in Fig. \ref{Figure:overallstructure}, the training of SUCode is divided into three stages. In Stage I, a set of semantic category-specific codebooks are learned from raw underwater images and their corresponding semantic masks, enabling fine-grained representation and addressing spatially inconsistent degradation problem. In Stage II, a weight-quantized coder is trained via a self-reconstruction task on raw underwater images, which aggregates the fixed semantic-aware codebooks into a unified representation, mitigating the ill-posed nature of UIE. In Stage III, this representation is used to train a dual-decoder  architecture specifically tailored for the enhance task. A feature fusion module adaptively combines decoded features from raw and enhanced images, facilitating robust domain transformation and feature modulation, generating visually coherent outputs.

\subsection{Semantic-aware Codebook Pretraining (Stage I)}

Underwater images exhibit diverse scenes and target differences, and various degradation factors can also affect different areas within a single image. Unlike approaches that encode a static one-shot codebook for all images \cite{wangCodeUNetAutonomousUnderwater2024} \cite{chenRealWorldBlindSuperResolution2022} or maintain and update a set of codebooks according the category the image level \cite{liuLearningImageAdaptiveCodebooks2023}, we aim to introduce codebook representation that distinguishes categories at the pixel level tailored to improve the generalization ability of UIE. Specifically, semantic-aware codebooks are learned guided by semantic masks, ensuring that each image region is encoded separately based on its semantic content (e.g., water, background, foreground objects) at the pixel level. The codebook is trained using the raw underwater images. This method not only diversifies the codebook but also distinguishes blur differences in different regions of the image based on semantic information.

\begin{figure}[h]
  \centering
  \includegraphics[width=\linewidth]{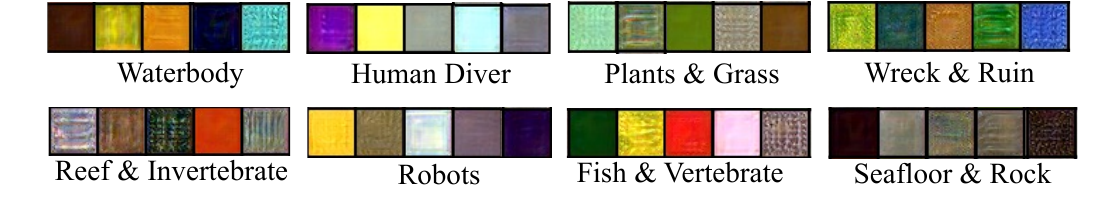}
  \caption{Visualization of the learned codebooks for different underwater categories. }
  \label{Figure:codebook}
\end{figure}

\begin{figure*}[h]
  \centering
  \includegraphics[width=0.9\linewidth]{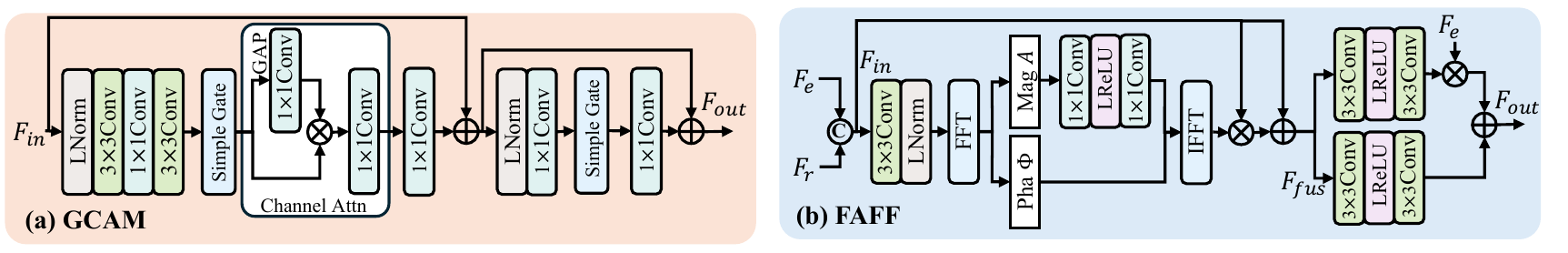}
  \caption{Structure of the proposed Gated Channel Attention Module (GCAM) and Frequency-Aware Feature Fusion (FAFF) module. rFFT and IrFFT represent real fast fourier transform and inverse real fast fourier transform, respecively. Mag and Pha represent the magnitude and phase variables, respectively. GAP refer to the global average pool operation.}
  \label{Figure:blocks}
\end{figure*}

\textbf{Codebook Quantization with Semantic Guidance.} Given an underwater image set with semantic annotations, we extend the standard quantized auto-encoder to learn class-specific codebooks by leveraging semantic segmentation masks during training. As shown in Fig. \ref{Figure:overallstructure}, the input underwater image $x \in \mathbb{R} ^{H \times W \times 3}$ is encoded by a multi-scale encoder $E_q$ to generate latent embeddings $\hat{z} = E_q(x) \in \mathbb{R} ^{h\times w \times n_z}$, where $n_z$ is the embedding dimension. To learn class-specific representations for underwater scenarios, we maintain a separate learnable codebooks set:
\begin{equation}
    \mathcal{Z} = \left\{Z_{c} \in \mathbb{R} ^{N \times n_z} \mid c = 1, 2, \dots, C \right\},
\end{equation}
where $N$ is the number of codes in the corresponding codebook, and $C$ is the number of semantic categories. $Z_{c}$ represents the codebook for category $c$. The semantic mask $m \in \mathbb{R} ^{H \times W }$ is resized to match the resolution of $\hat{z}$. Then, for each spatial location $i$, based on its class label $c(i)$, we find the discrete representation of $\hat{z}^{(i)}$ from the codebook corresponding to class $c(i)$. Specifically, the quantization process is achieved by minimizing the Euclidean distance between $\hat{z}^{(i)}$ and all the codebook vectors $Z_{c(i),j}$ in the codebook $Z_{c(i)}:$
\begin{equation}
    z_q^{(i)} = \mathop{\arg\min}\limits_{j \in \{0, \dots, N-1\}}\left \| \hat{z}^{(i)} -Z_{c(i),j}\right \|^2_2,
\end{equation}
where  $z_q^{(i)}$ denotes the quantized representation at location $i$, $\hat{z}^{(i)}$ is the original feature at location $i$, and $Z_{c(i),j}$ represent the $j$-th entry in codebook $Z_{c(i)}$ corresponding to class $c(i)$. All class-specific quantized outputs are then merged using the mask to form the final quantized embedding $\hat{z}_q$. Following the quantization process, the decoder $G_q$ reconstructs the high-resolution image $\hat{x}$ as: 
\begin{equation}
\hat{x}=G_q(\hat{z}_q)\approx x.
\end{equation}

We apply an adversarial learning scheme to train the encoder $E_q$, the codebook $Z$, and the decoder $G_q$ with a discriminator $D_q$. Compared to previous VQ-based method that quantize the entire image, our proposed method allows each semantic region to be encoded by a dedicated codebook, leading to improved restoration quality and class-specific representation learning. This is particularly advantageous for underwater scenes, where degradation is highly correlated with scene semantics. Some visual results of the codebooks are shown in Fig. \ref{Figure:codebook}.

\textbf{Loss Function.} The objective of stage I is to construct a discrete codebook representation using the raw underwater image and its semantic mask. To jointly training underwater image encoder $E_q$, codebook $\mathcal{Z}$, and decoder $G_q$, we adopt the pixel loss $\mathcal{L}_1$, the perceptual loss $\mathcal{L}_{per}$, the adversarial loss $\mathcal{L}_{adv}$, vector quantization loss $\mathcal{L}_{VQ}$:
\begin{equation}
\mathcal{L}_1 = \| \hat{x}-x\|_1
\end{equation}
\begin{equation}
   \mathcal{L}_{per} = \| \phi(\hat{x}) - \phi(x) \|_1
\end{equation}
\begin{equation}
      \mathcal{L}_{adv} = \log D(\hat{x}) + \log (1-D(x)).
\end{equation}
\begin{equation}
\begin{split}
\mathcal{L}_{VQ} = \| \mathrm{sg}[\hat{z}_q]-z_q \|_2^2 + \beta \| \hat{z}_q - \mathrm{sg}[z_q] \|_2^2 + \\ 
\lambda_s \| \mathrm{Conv}(\hat{z}_q) - \phi(x)\|_2^2, 
\end{split}
\end{equation}
where $\mathrm{sg}[\cdot]$ denotes the stop-gradient operation, $\mathrm{Conv}(\cdot)$ denotes a single convolutional layer to match the dimension of $\hat{z}_q$ and $\phi(x)$, $\phi(\cdot)$ denotes the VGG19 feature extractor, $\beta=0.25$ and $\lambda_s=0.1$ denotes the hyper-parameter, respectively. The total loss is expressed as:
\begin{equation}
    \mathcal{L}_{stage1} = \mathcal{L}_1 + \mathcal{L}_{per} + \lambda_{adv} \mathcal{L}_{adv} + \mathcal{L}_{VQ}(E_q,G_q),
\end{equation}
where $\lambda_{adv}=0.1$.

\subsection{Underwater Image Representation (Stage II)}
Traditional VQ-based image enhancement methods directly learn the mapping from raw degraded images to clear images with the help of discrete codebook representation. This approach is effective for natural images because it can learn discrete representations of true clear images. However, reference images in underwater environments are artificially selected pseudo ground truths. Directly using their feature mappings to clear images interferes with the network's ability to learn the true, accurate features of the underwater contents. To address this, we define the second stage of SUCode as the self-recovery stage, enabling the network to better recover the unique features of underwater images from the semantic-aware discrete representation extracted in the previous stage. 

In this stage, the raw underwater image is first mapped into class-specific quantized representations $\hat{z}_{q_c}$ using the encoder network $E_r$ and semantric-aware codebooks $Z_{c}$. A weighted predictor then synthesizes them to a  unified representation $\hat{z}_q$. Finally, a decoder $G_r$ composed by GCAMs reconstructs the raw image.

\textbf{Semantic-aware Codebook Partition and Synthesis.}
Given the semantic-aware codebooks $ \mathcal{Z} = \{Z_{c}$  $|  c = 1, 2, \dots, C \} $ pretrained in stage I, it is possible to map the image to a distinct discrete space in $C$ different ways. Each codebook $Z_{c}$ corresponds to a unique part of the underwater scenario in the feature space, allowing the model to learn semantic-specific representations. 

Specifically, given an underwater image $x$, $C$ quantized representations $ \mathcal{\hat{Z}}_q = \left\{\hat{z}_{q_c} \mid c = 1, 2, \dots, C \right\} $ are generated, with each representation $\hat{z}_{q_c}$ being a quantized representation obtained from the semantic codebook of category $c$. 

To recover underwater image from the $\mathcal{\hat{Z}}_q$, the representations must first be synthesized into a unified one $\hat{z}_q$.  This process is facilitated by a weight prediction module, which generates $C$ weight maps $\mathcal{W} =\left\{w_c \in \mathbb{R}^{h \times w \times 1} \mid c = 1, 2, \dots, C \right\}$. The weight predictor module consists of a Swin Transformer layer \cite{liuSwinTransformerHierarchical2021} and a convolution layer, which match the channels of weight map with $C$. The synthesis process of the quantized representations can thus be expressed as:

\begin{equation}
\hat{z}_q = \sum_c w_c \times \hat{z}_{q_c}.
\end{equation}

Finally, the adaptive feature $\hat{z}_q$ is reconstructed into the raw underwater image $\hat{x}$ using the decoder $G_r$.
\begin{equation}
\hat{x}=G_r(\hat{z}_q)\approx x.
\end{equation}

\textbf{Gated Channel Attention Module.}
Considering the varying attenuation of underwater light across wavelengths, underwater images suffer from severe color imbalance. Specifically, red light is absorbed first, causing the image to lean more towards blue and green. Therefore, the enhanced image is prone to over-amplification of colors. Furthermore, the illumination differences introduced by foreground objects and artificial light sources  \cite{liUIALNEnhancementUnderwater2023} may exacerbate these uneven color shift. To more effectively recover color details in underwater images from discrete representations, we introduce the GCAM, as shown in Fig. \ref{Figure:blocks}(a).
Specifically, GCAM achieves adaptive reweighting of color channels through a learnable gate-controlled attention mechanism. First, a simple gating operation \cite{feijooDarkIRRobustLowLight2025} and convolutions with global average pooling are applied to partition the channels and achieve lightweight channel attention, respectively. Then, a multi-layer perceptron with simple gating are used to synthesize the features. This allows the network to emphasize information-rich channels and suppress noise or oversaturated responses, resulting in more stable and realistic color restoration.

\textbf{Loss Function.}
The Stage II of SUCode is defined as the reconstruction stage of raw underwater images. To train the encoder $E_r$, the weight prediction module $\mathcal{W}$, and the decoder $G_r$, we employ four kinds of loss functions: pixel loss $\mathcal{L}_1$, perceptual loss $\mathcal{L}_{per}$, adversarial loss $\mathcal{L}_{adv}$, and vector quantization loss $\mathcal{L}_{VQ}$. The training objective in stage II is formulated as:
\begin{equation}
    \mathcal{L}_{stage2} = \mathcal{L}_1 + \mathcal{L}_{per} + \lambda_a \mathcal{L}_{adv} + \mathcal{L}_{VQ}(E_r,G_r),
\end{equation}
where $\lambda_{adv}=0.1$.

\subsection{Image Enhancement Learning (Stage III)}
Based on the discrete feature representations learned from the raw underwater images, SUCode formulates UIE as a domain-adaptive feature modulation problem. Specifically, with the vector quantizer and decoder $G_r$ kept frozen, we introduce FAFF modules that conditionally shift the raw-image features toward the feature space of the enhancement decoder $G_e$, producing the enhanced output $\hat{x}_{pred}$. Conceptually, this amounts to improving local appearance of underwater image by adaptively shifting the class-specific codebook while preserving high-level semantic content. Therefore, SUCode maintains semantic understanding of the scene and adaptively handles diverse underwater degradation patterns, mitigating the visual artifacts and instability that commonly affect traditional UIE methods.

\textbf{Frequency-Aware Feature Fusion.}
To bridge the domain gap between raw and enhanced underwater image representations, the decoder $G_e$ integrates the FAFF, which explicitly fuses complementary cues from both the raw and enhanced decoding streams, while maintaining the structural prior provided by $G_r$. The structure of FAFF is shown in Fig. \ref{Figure:blocks}(b). The key idea of FAFF is to retain structural fidelity from the raw domain while adaptively infusing rich texture and color details from the enhanced domain. Concretely, we preserve the phase of the original features to maintain structure, selectively amplify the magnitude guided by the enhanced features, and apply domain-adaptive modulation via a learnable affine transformation. Operating in the spectral domain allows the method to inject enhanced appearance information into the output while locking the original shape cues, thereby preserving the contours and semantics encoded by the codebook despite the domain gap.

Specifically, given the raw-domain features $F_r \in \mathbb{R}^{H_i \times W_i \times C_i}$ and the enhance-domain features $F_e \in \mathbb{R}^{H_i \times W_i \times C_i}$, FAFF first concatenates them along channels to obtained $F_{in}$ and apply a $3 \times 3$ convolution followed by layer normalization to an aligned representation. To exploit spectral cues while preserving geometric structure, a channel-wise real Fast Fourier Transform (rFFT) \cite{feijooDarkIRRobustLowLight2025} is performed to obtain a one-sided spectrum:
\begin{equation}
    \hat{F} = \mathrm{rFFT}_2(F_{in}).
\end{equation}
Subsequently, phase-amplitude decomposition is performed to obtain magnitude $A=|\hat{F}|$ and $\Phi = \angle \hat{F}$. A lightweight frequency mapper $g_\theta$ is then applied to the magnitude only, which includes $1 \times 1$ convolutions and LeakyReLU operation, yielding $A'=g_\theta(A)$. Keeping the original phase $\Phi$ for structural fidelity, the spectrum is reconstructed as $\hat{F'}=g_\theta(A)\odot e^{j\Phi}$ and transformed back to the spatial domain:
\begin{equation}
    F_{freq}=\mathrm{=IrFFT}_2(\hat{F'}, s=(H_i,W_i)).
\end{equation}
This frequency-aware feature acts as a self-modulation mask on $F_{in}$:
\begin{equation}
F_{fus}=F_{in} + \gamma(F_{in} \cdot F_{freq}),
 \end{equation}
where $\gamma$ is a learnable per-channel coefficient. Finally, two parallel spatial branches $\psi_{scale}$ and $\psi_{shift}$ are used to generate an adaptive scale-shift modulate the output feature $F_{out} \in \mathbb{R}^{H_i \times W_i \times C_i}$:
\begin{equation}
    F_{out} = \psi_{scale}(F_{fus}) \cdot F_{e} + \psi_{shift}(F_{fus}).
\end{equation}
The parallel spatial branches are constructed with $3 \times 3$ convolutions and LeakyReLU operation. In summary, this design selectively enhances texture and lighting through amplitude reweighting while preserving spatial structure through phase consistency.


\textbf{Loss Function.}
In stage III, adaptive image enhancement is achieved from the discrete representation of the image in the Codebook and the decoder features of the raw image. To train the encoder $E_r$ and the decoder $G_e$, we use four types of loss functions: pixel loss $\mathcal{L}'_1$, perceptual loss $\mathcal{L}'_{per}$, adversarial loss $\mathcal{L}'_{adv}$, and code-level loss $\mathcal{L}'_{code}$, given as:
\begin{equation}
\mathcal{L}_1' = \| \hat{x}_{pred}-x_{gt}\|_1
\end{equation}
\begin{equation}
   \mathcal{L}'_{per} = \| \phi(\hat{x}_{pred}) - \phi(x_{gt}) \|_1
\end{equation}
\begin{equation}
      \mathcal{L}'_{adv} = \log D(\hat{x}_{pred}) + \log (1-D(x_{gt})).
\end{equation}
where $x_{gt}$ refer to the ground truth image. The code loss \cite{liuLearningImageAdaptiveCodebooks2023}\cite{chenSimpleFrameworkContrastive2020} is to reduce the difference between the discrete representation $z_{gt}$ of the ground truth and representation $\hat{z}$:
\begin{equation}
\mathcal{L}_{code} = \beta \left \| \hat{z}-\mathrm{sg}[z_{gt}] \right \|^2_2.
\end{equation}
where $\mathrm{sg}$ is the stop-gradient operation and $\beta$ is set to $0.25$ to control the codebook update frequency. $z_{gt}$ is obtained by $E_r$ and the vector quantizer. The total loss is expressed as:
\begin{equation}
    \mathcal{L}_{stage3} = \mathcal{L}'_1 + \mathcal{L}'_{per} + \lambda_{adv}' \mathcal{L}'_{adv} + \mathcal{L}_{code},
\end{equation}
where $\lambda_{adv}'=0.1$.

\section{Experiments and Results}\label{Section:4}
In this section, we first introduce the experimental settings of the proposed SUCode, followed by qualitative and quantitative comparisons with other state-of-the-art UIE methods. Then, we conduct ablation study to verify the modules introduced in SUCode. Finally, we verify the effectiveness of SUCode on downstream tasks. 

\begin{table*}[ht!]\centering
\caption{Quantitative comparison of different UIE methods on the UIEB \cite{liUnderwaterImageEnhancement2020} and SUIM-E \cite{qiSGUIENetSemanticAttention2022} datasets. The best results are highlighted in \textbf{bold} and the second best results are \underline{underlined}.}\label{Table:baselineresult}
\scalebox{0.85}{
\begin{tabular}{c|ccccc|ccccc}
\hline\toprule
\multirow{2}{*}{\textbf{Method}} & \multicolumn{5}{c|}{\textbf{SUIM-E\cite{qiSGUIENetSemanticAttention2022}}} & \multicolumn{5}{c}{\textbf{UIEB\cite{liUnderwaterImageEnhancement2020}}} \\
\cmidrule{2-11} 
& SSIM$\uparrow$ & PSNR$\uparrow$ & LPIPS$\downarrow$ & UCIQE$\uparrow$ & UIQM$\uparrow$
& SSIM$\uparrow$ & PSNR$\uparrow$ & LPIPS$\downarrow$ & UCIQE$\uparrow$ & UIQM$\uparrow$ \\
\midrule
Fusion\cite{ancutiEnhancingUnderwaterImages2012}   & 0.876 & 16.824 & 0.226 & 58.413 & 2.811 & 0.907 & 18.483 & 0.211 & 52.823 & 3.251 \\
IBLA\cite{pengUnderwaterImageRestoration2017a}     & 0.788 & 16.019 & 0.221 & 62.498 & 1.870 & 0.771 & 15.009 & 0.341 & 53.816 & 2.346  \\
ULAP\cite{songEnhancementUnderwaterImages2020}    & 0.860 & 16.574 & 0.232 & 59.746 & 2.174 & 0.902 & 17.871 & 0.233 & 52.620 & 3.309 \\
UDCP\cite{houNovelDarkChannel2020}     & 0.581 & 11.694 & 0.308 & 62.172 & 1.815 & 0.603 & 11.001 & 0.399 & 59.492 & 2.147 \\
\midrule
WaterNet\cite{liUnderwaterImageEnhancement2020} & 0.907 & 22.295 & 0.144 & 60.999 & 2.807 & 0.898 & 21.566 & 0.237 & 61.805 & 3.314 \\
UColor\cite{liUnderwaterImageEnhancement2021}   & 0.898 & 22.860 & 0.145 & 62.436 & 2.860 & 0.906 & 22.266 & 0.187 & 59.176 & 3.316  \\
UShape\cite{pengUShapeTransformerUnderwater2023}   & 0.851 & 21.369 & 0.147 & 53.451 & \textbf{2.969} & 0.819 & 20.266 & 0.219 & 48.406 & 3.296 \\
CCMSR\cite{qiDeepColorCorrectedMultiscale2024a} & 0.896 & 22.028 & 0.161 & 60.129 & 2.875 & 0.914 & 22.761 & 0.180 & 57.084 & 3.274  \\
WfDiff\cite{zhaoWaveletbasedFourierInformation2024}  & 0.853 & 16.176 & 0.184 & 57.052 & 2.701 & 0.888 & 18.994 & 0.214 & 53.269 & 3.255 \\
SMDR-IS\cite{zhangSynergisticMultiscaleDetail2024}  & 0.896 & 22.082 & 0.146 & 62.600 & 2.749 & \underline{0.924} & 22.232 & 0.166 & 61.559 & 2.952 \\
AMSIN\cite{quanEnhancingUnderwaterImages2024a}    & 0.902 & 21.923 & 0.125 & 61.399 & 2.762 & 0.921 & 22.635 & 0.146 & 62.332 & 3.309 \\
HCLR-Net\cite{quanEnhancingUnderwaterImages2024a}    & 0.914 & 20.749 & 0.185 & 58.765 & 3.360 & 0.902 & 22.317 & \textbf{0.124} & 58.599 & 3.279 \\
RUE-Net\cite{wangRUENetAdvancingUnderwater2024}  & \underline{0.921} & \underline{22.902} & 0.121 & 62.500 & 2.776 & 0.923 & 22.743 & 0.164 & 62.357 & 3.260  \\
FDCE-Net\cite{chengFDCENetUnderwaterImage2025a}  & 0.912 & 22.840 & 0.120 & 60.722 & 2.872 & 0.923 & \underline{23.039} & 0.141 & 58.765 & \underline{3.360}  \\
SS-UIE\cite{pengAdaptiveDualdomainLearning2025}   & 0.871 & 21.713 & 0.182 & 59.538 & 2.815 & 0.850 & 21.006 & 0.255 & 58.919 & 3.066  \\
CDF-UIE\cite{zhangCDFUIELeveragingCrossDomain2025a}  & 0.892 & 22.089 & 0.116 & 54.826 & 2.838 & 0.886 & 21.592 & 0.159 & 54.219 & 3.333  \\
\midrule
FeMaSR\cite{chenRealWorldBlindSuperResolution2022}  & 0.908 & 22.749 & \underline{0.100} & \underline{62.605} & 2.841 & 0.883 & 22.733 & \underline{0.137} & \underline{62.675} & 3.301  \\
AdaCode\cite{liuLearningImageAdaptiveCodebooks2023}  & 0.886 & 22.329 & 0.105 & 62.409 & 2.812  & 0.818 & 21.792 & 0.156 & 60.835 & 3.216   \\
RIDCP\cite{wuRIDCPRevitalizingReal2023a}& 0.509 & 13.407 & 0.572 & 42.184 & 2.533 & 0.573 & 14.915 & 0.487 & 48.679 & 2.246  \\
IPC-Dehaze\cite{fuIterativePredictorCriticCode2025}& 0.823 & 13.869 & 0.381 & 50.837 & 2.252 & 0.852 & 16.923 & 0.226 & 54.777 & 2.352  \\
CodeUNet\cite{wangCodeUNetAutonomousUnderwater2024}& 0.590 & 17.349 & 0.447 & 54.769 & 2.705 & 0.836 & 21.468 & 0.196 & 59.650 & \textbf{3.383}  \\
\midrule
SUCode(Ours)     & \textbf{0.939} & \textbf{23.908} & \textbf{0.087} & \textbf{62.618} & \underline{2.878} & \textbf{0.925} & \textbf{23.857} & \textbf{0.124} & \textbf{63.136} & 3.174 \\
\bottomrule
\end{tabular}}
\end{table*}

\begin{figure*}[h]
  \centering
  \includegraphics[width=0.75\linewidth]{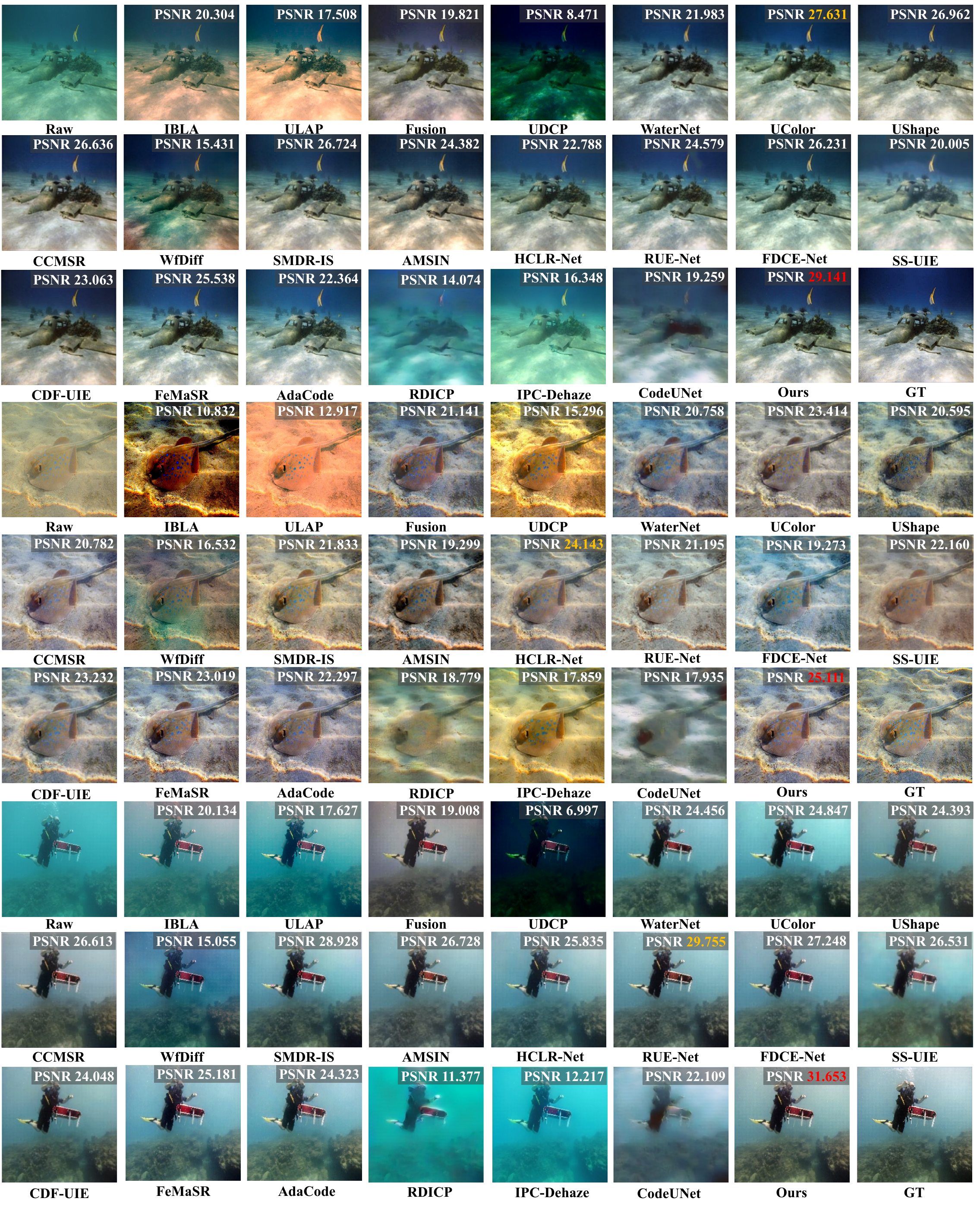}
  \caption{Visual comparison of UIE results sampled from the test set of SUIM-E \cite{qiSGUIENetSemanticAttention2022} dataset. The highest PSNR results are highlighted in red and the second best results are yellow.}
  \label{Figure:suim}
\end{figure*}

\begin{figure*}[h]
  \centering
  \includegraphics[width=0.75\linewidth]{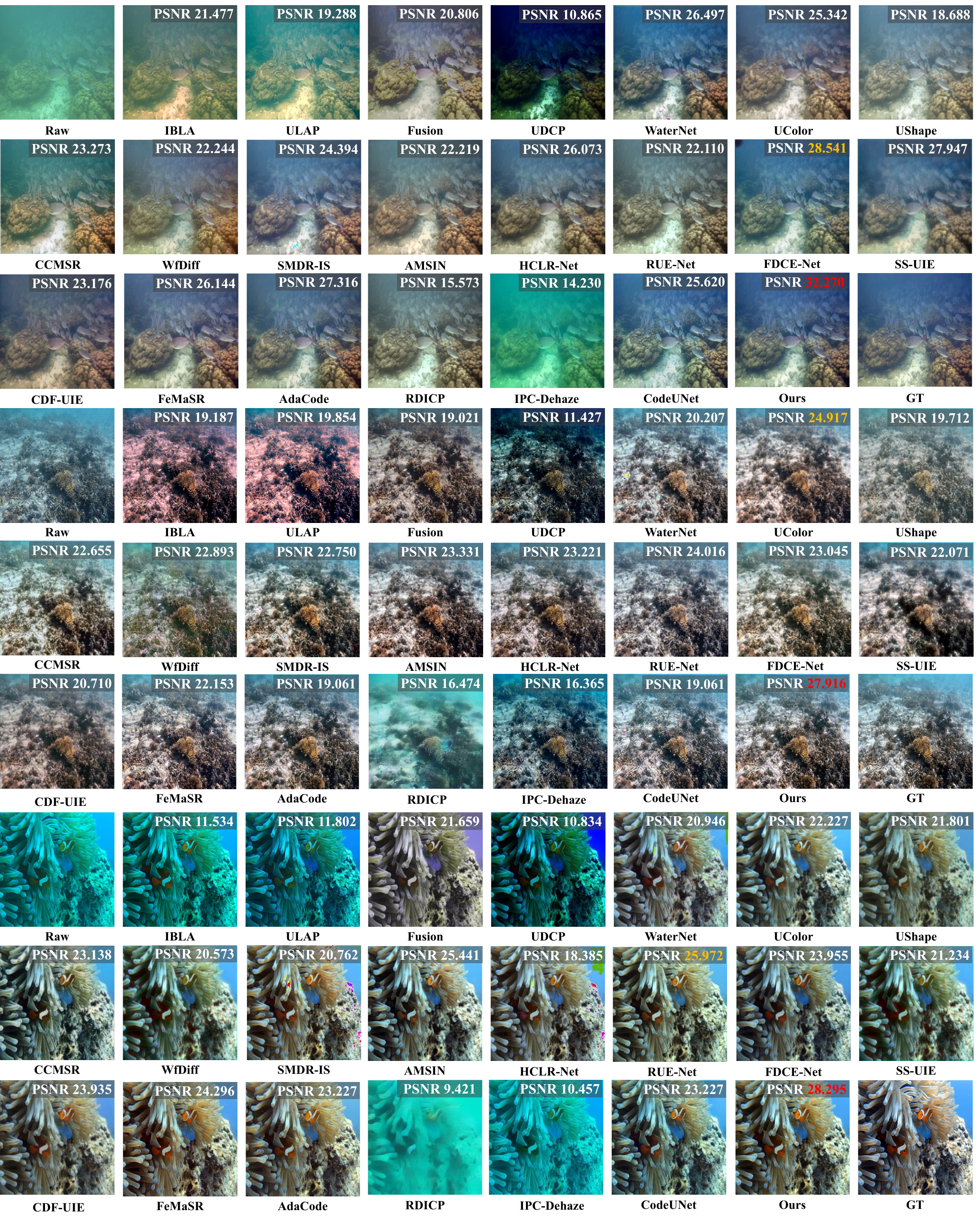}
  \caption{Visual comparison of UIE results sampled from the test set of UIEB \cite{liUnderwaterImageEnhancement2020} dataset. The highest PSNR results are highlighted in red and the second best results are yellow.}
  \label{Figure:uieb}
\end{figure*}

\subsection{Experimental Settings}

\textbf{Datasets and Training Details.}
We trained SUCode using the SUIM-E \cite{qiSGUIENetSemanticAttention2022} and UIEB \cite{liUnderwaterImageEnhancement2020} datasets. The SUIM-E dataset contains 1635 images with segmentation masks and enhanced ground truth, where 1525 images are used for training and 110 for testing. The UIEB dataset has 890 images, from which 800 are used for training and 90 for testing.

SUCode is trained in three stages. In Stage I, we train using raw images and segmentation masks in SUIM-E dataset to obtain semantic codebooks and restore the raw underwater images. The underwater scenes is categorized into 8 classes \cite{islamSemanticSegmentationUnderwater2020}: human divers, aquatic plants, wrecks, underwater robots, reefs \& invertebrates, fish \& vertebrates, sea-floor \& rocks, and water body. These categories correspond to 8 codebooks for Stage I. In Stages II and III, we use the trained codebooks to train and test on the SUIM-E and UIEB datasets, respectively, to enhance the underwater images. The codebook size is set to $256 \times 256$, and the input image is represented as a $32 \times 32$ code sequence. 

To assess the robustness of SUCode, we further perform cross-dataset evaluation on the LSUI \cite{pengUShapeTransformerUnderwater2023} and UFO120 \cite{islamSimultaneousEnhancementSuperresolution2020a} datasets using the pretrained model on UIEB. Among them, we selected 400 images and 120 images from the LSUI dataset UFO120 dataset for evaluation, respectively.

\textbf{Implementation Details.}
SUCode is implemented in PyTorch and trained on an NVIDIA 3090Ti GPU. The training consists of 200 epochs with a batch size of 4, using the Adam optimizer. The learning rates for the generator and discriminator are set to 1e-4 and 4e-4, respectively. Training and testing are performed at $256 \times 256$ resolution, with random cropping during training and direct scaling during testing.

\begin{table*}[ht!]\centering
\caption{Cross dataset quantitative comparison of different UIE methods trained on the UIEB dataset and evaluate on the LSUI \cite{pengUShapeTransformerUnderwater2023} and UFO-120 \cite{islamSimultaneousEnhancementSuperresolution2020a} datasets. The best results are highlighted in \textbf{bold} and the second best results are \underline{underlined}.}\label{Table:crossresult}
\scalebox{0.85}{
\begin{tabular}{c|ccccc|ccccc}
\hline\toprule
\multirow{2}{*}{\textbf{Method}} & \multicolumn{5}{c|}{\textbf{LSUI \cite{pengUShapeTransformerUnderwater2023}}} & \multicolumn{5}{c}{\textbf{UFO-120 \cite{islamSimultaneousEnhancementSuperresolution2020a}}} \\
\cmidrule{2-11} 
& SSIM$\uparrow$ & PSNR$\uparrow$ & LPIPS$\downarrow$ & UCIQE$\uparrow$ & UIQM$\uparrow$
& SSIM$\uparrow$ & PSNR$\uparrow$ & LPIPS$\downarrow$ & UCIQE$\uparrow$ & UIQM$\uparrow$ \\
\midrule
WaterNet\cite{liUnderwaterImageEnhancement2020} & 0.810 & 19.024 & 0.342 & 63.555 & 3.207 & \underline{0.833} & 18.424 & 0.271 & 62.882 & 3.130  \\
UColor\cite{liUnderwaterImageEnhancement2021}   & 0.821 & 20.017 & 0.300 & 62.016 & 3.195 & 0.788 & 18.355 & 0.360 & 63.363 & 3.137  \\
UShape\cite{pengUShapeTransformerUnderwater2023}   & 0.766 & 18.604 & 0.320 & 50.992 & \underline{3.264} & 0.789 & 18.624 & 0.321 & 57.875 & \textbf{3.218}  \\
CCMSR\cite{qiDeepColorCorrectedMultiscale2024a} & 0.839 & \textbf{20.678} & 0.282 & 60.024 & 3.180 & 0.802 & 18.840 & 0.331 & 60.667 & 3.148 \\
WfDiff\cite{zhaoWaveletbasedFourierInformation2024}  & \underline{0.855} & 19.400 & 0.276 & 59.013 & 3.121 & 0.797 & \underline{19.675} & 0.319 & 62.320 & 3.054  \\
SMDR-IS\cite{zhangSynergisticMultiscaleDetail2024}  & 0.815 & 19.246 & 0.304 & 62.617 & 3.098 & 0.780 & 17.612 & 0.343 & 64.184 & 3.020  \\
AMSIN\cite{quanEnhancingUnderwaterImages2024a}    & 0.826 & 19.792 & \underline{0.260} & \underline{64.288} & 3.100 & 0.786 & 17.822 & 0.314 & \textbf{65.704} & 3.058  \\
HCLR-Net\cite{quanEnhancingUnderwaterImages2024a}    & 0.821 & 18.897 & 0.312 & 61.274 & 3.059 & 0.801 & 18.462 & 0.336 & 62.769 & 3.081 \\
RUE-Net\cite{wangRUENetAdvancingUnderwater2024}  & 0.854 & 20.651 & 0.261 & 64.076 & 3.089 & 0.817 & 18.920 & 0.312 & 64.249 & 3.024  \\
FDCE-Net\cite{chengFDCENetUnderwaterImage2025a}  & 0.811 & 18.649 & 0.327 & 62.386 & 3.193 & 0.827 & \textbf{20.556} & \underline{0.266} & 63.428 & 3.165  \\
SS-UIE\cite{pengAdaptiveDualdomainLearning2025}   & 0.821 & 19.708 & 0.331 & 60.552 & 3.106 & 0.785 & 18.565 & 0.343 & 65.097 & 2.794 \\
CodeUNet\cite{wangCodeUNetAutonomousUnderwater2024}& 0.804 & 20.040 & 0.263 & 62.467 & \textbf{3.269} & 0.783 & 18.935 & 0.285 & 62.317 & \underline{3.177}  \\
\midrule
SUCode(Ours)     & \textbf{0.860} & \underline{20.662} & \textbf{0.241} & \textbf{64.361} & 3.044 & \textbf{0.835} & 19.035 & \textbf{0.259} & \underline{65.635} & 2.936  \\
\bottomrule
\end{tabular}}
\end{table*}

\begin{figure*}[h]
  \centering
  \includegraphics[width=0.9\linewidth]{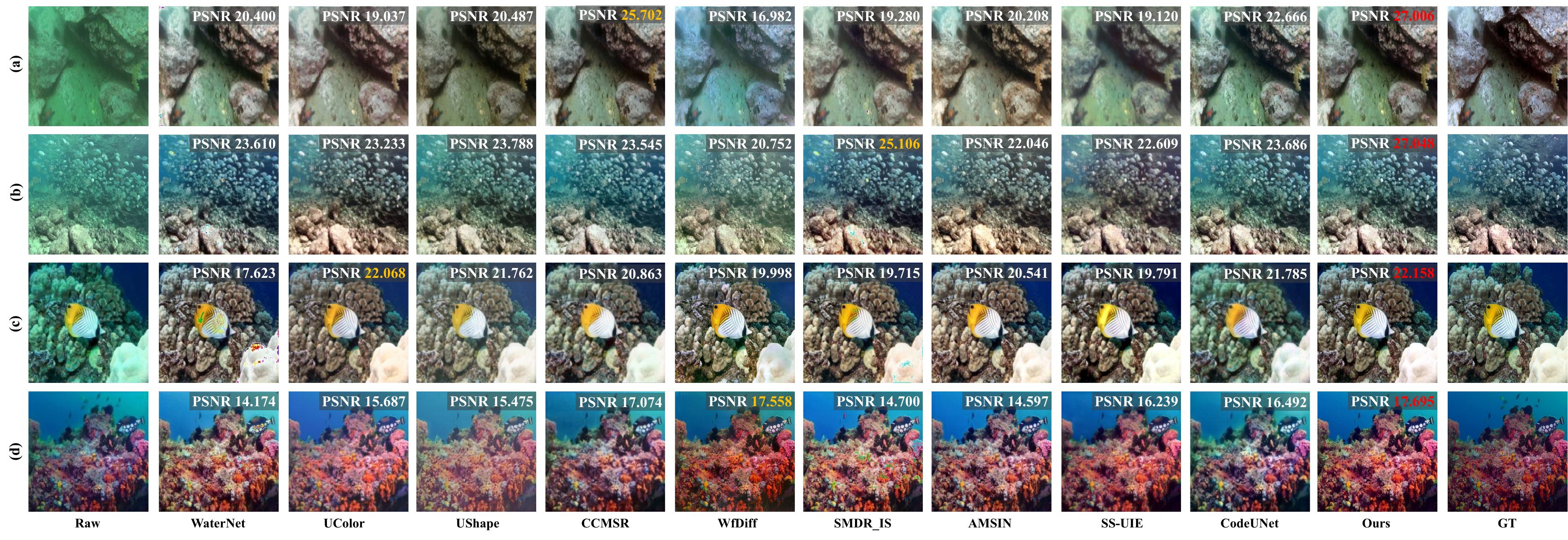}
  \caption{Visual comparison of cross-dataset evaluation results sampled from LSUI\cite{pengUShapeTransformerUnderwater2023} (a,b) and UFO120\cite{islamSimultaneousEnhancementSuperresolution2020a} (c,d) dataset. The highest PSNR results are highlighted in red and the second best results are yellow.}
  \label{Figure:cross}
\end{figure*}

\textbf{Evaluation Metrics.}
We evaluate SUCode using both reference and no-reference metrics. Reference metrics include SSIM \cite{wangImageQualityAssessment2004a}, PSNR \cite{korhonenPeakSignaltonoiseRatio2012}, and LPIPS \cite{zhangUnreasonableEffectivenessDeep2018}, which are used to evaluate the similarity between the predicted images and the ground-truth. The no-reference metrics include UCIQE \cite{zhangUnreasonableEffectivenessDeep2018} and UIQM \cite{zhangUnreasonableEffectivenessDeep2018}, which evaluate the color, clarity, and contrast from different levels.

\textbf{Comparison Methods.}
We compare SUCode with both traditional UIE methods and deep-learning-based UIE methods. The traditional UIE methods are compared including Fusion \cite{ancutiEnhancingUnderwaterImages2012}, IBLA\cite{pengUnderwaterImageRestoration2017a}, ULAP\cite{songEnhancementUnderwaterImages2020}, and UDCP\cite{houNovelDarkChannel2020}. The state-of-the-art deep learning-based UIE methods are compared including WaterNet \cite{liUnderwaterImageEnhancement2020}, UColor \cite{liUnderwaterImageEnhancement2021}, UShape \cite{pengUShapeTransformerUnderwater2023}, CCMSR \cite{qiDeepColorCorrectedMultiscale2024a}, WfDiff \cite{zhaoWaveletbasedFourierInformation2024}, SMDR-IS \cite{zhangSynergisticMultiscaleDetail2024}, HCLR-Net\cite{quanEnhancingUnderwaterImages2024a}, AMSIN \cite{quanEnhancingUnderwaterImages2024a}, RUE-Net\cite{wangRUENetAdvancingUnderwater2024}, SS-UIE \cite{pengAdaptiveDualdomainLearning2025}, FDCE-Net\cite{chengFDCENetUnderwaterImage2025a} and CDF-UIE\cite{zhangCDFUIELeveragingCrossDomain2025a}. We further compared the proposed SUCode with other Codebook-based methods, including FeMaSR \cite{chenRealWorldBlindSuperResolution2022}, AdaCode \cite{liuLearningImageAdaptiveCodebooks2023}, RIDCP\cite{wuRIDCPRevitalizingReal2023a}, IPC-Dehaze\cite{fuIterativePredictorCriticCode2025} and CodeUNet \cite{wangCodeUNetAutonomousUnderwater2024}. We retrain these models using the source codes from the respective authors and follow their experimental settings.

\subsection{Benchmarking Comparison Results}

\textbf{Quantitative Evaluation.}
We conducted training on the UIEB and SUIM-E datasets and perform quantitative comparisons using full-reference and no-reference evaluations to demonstrate the effectiveness of the proposed SUCode. The statistical results are shown in Table \ref{Table:baselineresult}. 

As shown in Table \ref{Table:baselineresult}, the proposed SUCode achieves the best performance across all full-reference metrics, including PSNR, SSIM and LPIPS. Compared to traditional UIE methods such as Fusion \cite{ancutiEnhancingUnderwaterImages2012} and ULAP \cite{songEnhancementUnderwaterImages2020}, which lack adaptability to complex underwater environments and objects, SUCode significantly outperforms them. Regarding deep learning methods, CCMSR \cite{qiDeepColorCorrectedMultiscale2024a} incorporates a physical model of underwater imaging, UColor \cite{liUnderwaterImageEnhancement2021} introduces additional image depth information, and SMDR-IS \cite{zhangSynergisticMultiscaleDetail2024} introduce explicit and implicit multi-scale image processing strategies, respectively. However, these methods treat the image as a whole and fail to leverage semantic information to differential enhance the foreground and background, resulting in inferior performance compared to the proposed SUCode method. FDCE-Net \cite{chengFDCENetUnderwaterImage2025a} performs enhancement operations in the frequency domain. however, this method lacks modeling of foreground and background, resulting in limited improvement in SSIM and LPIPS values. When compared with other Codebook-based methods, the proposed SUCode also achieved higher indicators. It should be noted that AdaCode \cite{liuLearningImageAdaptiveCodebooks2023} did not perform well in the UIE task, which may be because the complexity of underwater images makes constructing a Codebook according to the overall image category ineffective. 

We also report the no-reference metrics on SUIM-E and UIEB datasets in Table \ref{Table:baselineresult}, which shown the proposed SUCode performs comparably to, or better than, other UIE methods on these metrics. Specifically, the proposed SUCode achieves the highest UCIQE value on both SUIM-E and UIEB datasets, and the second highest UIQM value on SUIM-E. It is important to note that due to the way non-reference underwater image quality metrics are evaluated, the UIQM metrics are more focused on color and contrast. Specifically, images with higher color saturation, contrast difference, and brightness receive higher scores. These metrics don't consider the semantic information of the image and therefore often differ from human visual perception \cite{qiDeepColorCorrectedMultiscale2024a} \cite{zhangCDFUIELeveragingCrossDomain2025a} \cite{liUnderwaterImageEnhancement2021}. Therefore, non-reference metrics need to be combined with full-reference ones to evaluate the UIE model.

\textbf{Qualitative Evaluation.}
We conducted visual comparisons for qualitative evaluation between the proposed SUCode and other UIE methods, as is shown in Fig. \ref{Figure:suim} and Fig. \ref{Figure:uieb}. The visual comparisons  reveal that SUCode's enhancement results closely match the reference image, with more realistic colors, clearer local details, and fewer artifacts or unnatural areas. Specifically, SUCode effectively balances the foreground target and background, resulting in more realistic color restoration and producing images that align more closely with human visual perception. In contrast, CCMSR \cite{qiDeepColorCorrectedMultiscale2024a} enhance the image as a whole, which can introduce unnatural colors when the foreground and background differ significantly. Further, SUCode can restore images with minimal artifacts, which is crucial for downstream applications, where visual clarity is paramount. These results highlight the effectiveness of SUCode in restoring images by leveraging semantic information to model category-specific discrete codebooks. The SS-UIE \cite{pengAdaptiveDualdomainLearning2025} and WfDiff \cite{zhaoWaveletbasedFourierInformation2024} exhibit limitations in recovering complex textured regions, as these methods clearly lack clarity in background regions of the image. Additionally, other codebook-based methods \cite{chenRealWorldBlindSuperResolution2022} \cite{liuLearningImageAdaptiveCodebooks2023}  are more likely to introduce artifacts and unnatural details because they learn image features from pseudo ground-truth values that can be adversely affected. Methods specifically designed for natural images, such as RIDCP\cite{wuRIDCPRevitalizingReal2023a} and IPC-Dehaze\cite{fuIterativePredictorCriticCode2025}, exhibit poor adaptability to diverse underwater scenes and fail to produce consistent enhancement results. However, the proposed SUCode makes advantages of the FAFF, which can integrate features from the $G_r$ and $G_e$ decoders, making better use of discrete representations and enhancing the image quality.

\begin{table*}[]\centering
\caption{Comparison of different UIE methods in terms of number of parameters and FLOPs.}\label{Table:efficiency}
\setlength{\tabcolsep}{4.5pt} 
\scalebox{0.85}{
\begin{tabular}{c|ccccccccc|c}
\toprule
 Metric  & \makecell[c]{WaterNet\\ \cite{liUnderwaterImageEnhancement2020}} & \makecell[c]{UColor\\ \cite{liUnderwaterImageEnhancement2021}} & \makecell[c]{UShape\\ \cite{pengUShapeTransformerUnderwater2023}} & \makecell[c]{CCMSR\\ \cite{qiDeepColorCorrectedMultiscale2024a}} & \makecell[c]{WfDiff\\ \cite{zhaoWaveletbasedFourierInformation2024}} & \makecell[c]{SMDR-IS\\ \cite{zhangSynergisticMultiscaleDetail2024}} & \makecell[c]{AMSIN\\ \cite{quanEnhancingUnderwaterImages2024a}} & \makecell[c]{SS-UIE\\ \cite{pengAdaptiveDualdomainLearning2025}} & \makecell[c]{CodeUNet\\ \cite{wangCodeUNetAutonomousUnderwater2024}} & SUCode(Ours) \\ 
 \midrule
\#Params (M)& 1.09 & 148.77 & 31.58 & 21.13 & 100.55 & 29.33 & 4.67 & 20.63  & 29.47 & 61.18  \\
FLOPs (G) & 71.48 & 1402.18 & 0.86 & 43.60 & 369.29 & 46.27 & 37.28 & 40.09 & 111.49 & 133.31  \\ 
\bottomrule
\end{tabular}}
\end{table*}

\subsection{Cross-Dataset Evaluation Results.}
To further evaluate the robustness of the proposed method, we conduct cross-dataset validation experiments. Specifically, we use the SUCode model trained on the UIEB dataset and test it on the validation sets of the LSUI and UFO120 datasets, respectively. The statistical results and visual comparisons are shown in Table \ref{Table:crossresult}  and Fig. \ref{Figure:cross}. Statistical results show that this pre-trained SUCode achieved the best results in SSIM and LPIPS values on both datasets, and achieved competitive PSNR and UCIQE values, demonstrating superior generalization performance. This can be attributed to the advantage of the semantic-aware codebooks obtained by SUCode in the first stage: modeling category-related underwater image features can enable the UIE network to utilize target information from the raw underwater images to improve the enhanced adaptability. As shown in the visual comparison, other UIE methods fail to handle the foreground and background effectively, often introducing poor textures, over-smoothed details, or undesirable color shifts. In contrast, SUCode consistently selects appropriate enhancement regions for different target regions, restoring more natural color, brightness, and contrast. This is made possible by the cross-channel color restoration capability of the GCAM module, which helps the network recover a clear image from a discrete codebook based on a gated channel attention module.

\subsection{Efficiency Evaluation}
Table \ref{Table:efficiency} lists the analysis of network efficiency, expressed in terms of model size and computational complexity, which shows that SUCode achieves the moderate space complexity and time complexity. It should be noted that WaterNet \cite{liUnderwaterImageEnhancement2020} and UColor \cite{liUnderwaterImageEnhancement2021} require additional image preprocessing processing for enhancement, while the proposed SUCode can achieve end-to-end training and inference processes.

\begin{figure}[h]
  \centering
  \includegraphics[width=0.8\linewidth]{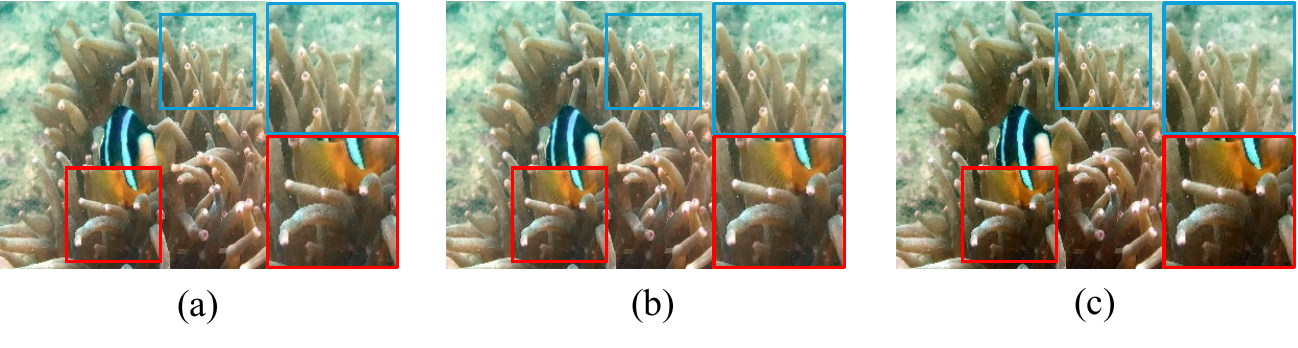}
  \caption{Visual comparison of ablation studies on the codebook quantization strategy. (a) One shot codebook quantization. (b) Image category-based codebook quantization. (c) SUCode. }
  \label{Figure:ab_codebook}
\end{figure}

\begin{table}[]\centering
\caption{Ablation study of the effectiveness of codebook generation methods. The best results are highlighted in \textbf{bold}}\label{Table:codebook}
\scalebox{0.8}{
\begin{tabular}{ccc|ccc}
\toprule
 OS    & ICC   & Ours   & PSNR$\uparrow$ & SSIM$\uparrow$ & LPIPS$\downarrow$ \\
\midrule
$\surd$&       &       & 23.095 & 0.917 & 0.142 \\
       &$\surd$&       & 23.245 & 0.916 & 0.134 \\
       &       &$\surd$& \textbf{23.857} & \textbf{0.925} & \textbf{0.124}\\
\bottomrule
\end{tabular}}
\end{table} 
\subsection{Ablation Study}
We conducted ablation studies on the UIEB dataset to evaluate the effectiveness of each component in SUCode.

\textbf{Effectiveness of Semantic-Aware Codebook.}
To analyze the effectiveness of the proposed semantic-aware codebook, we perform ablation studies on the codebook quantization strategy. The one shot (OS) quantization methods quantizes a single and fixed codebook for all images \cite{esserTamingTransformersHighResolution2021}\cite{chenRealWorldBlindSuperResolution2022}, while the image category-based codebook (ICC) quantization methods learns a set of foundational codebooks that correspond to different image categories, and then dynamically combines them based on the input image's characteristics. The enhancement process comparison of these methods is shown in Fig. \ref{Figure:introduction}. It can be seen from the visual comparison in Fig. \ref{Figure:ab_codebook} that the proposed SUCode introduces more reasonable contrast and clarity in the near coral regions and alleviates the blue-green color cast in the distant background regions compared to the SC and ICC methods.  The quantitative results are summarized in Table \ref{Table:codebook}, in which it is observed that the proposed semantic-aware codebook quantization strategy provides reasonable performance, which leveraging pixel-wise codebook updates guided by semantic masks.  This is mainly due to the complexity and diversity of the underwater environment. Therefore, the codebook established on the fine-grained area can more completely represent the image.

\begin{figure}[h]
  \centering
  \includegraphics[width=0.8\linewidth]{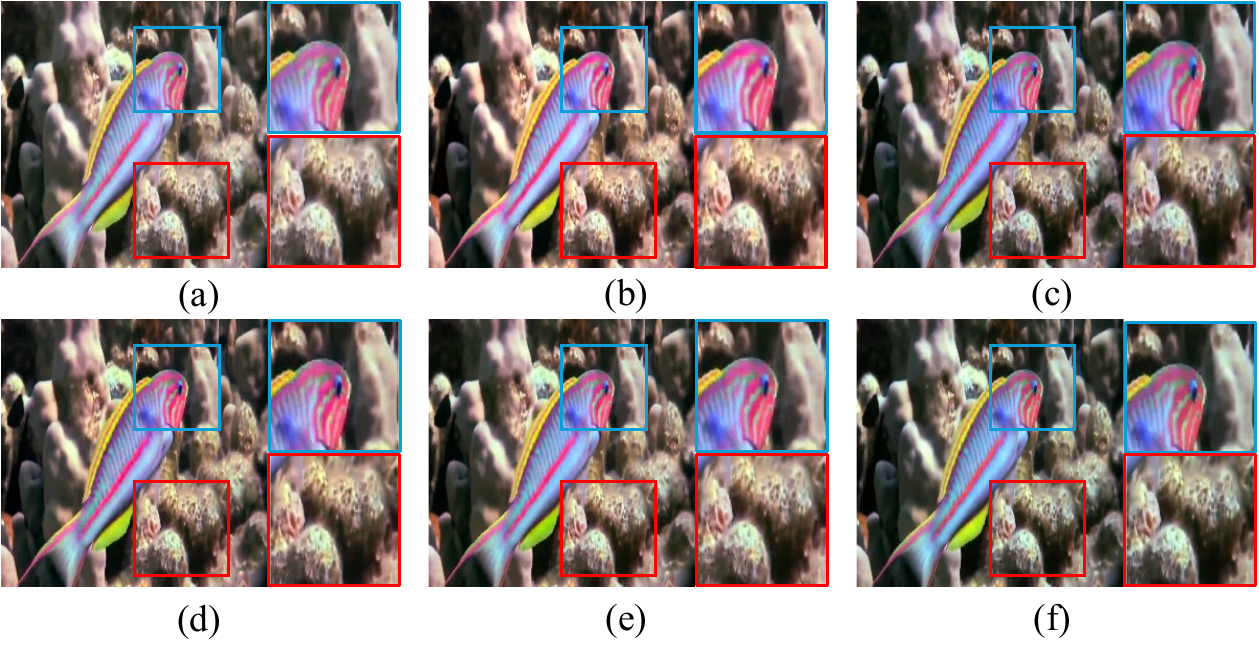}
  \caption{Visual comparison of ablation studies on the network modules. (a) Baseline model. (b) Model with FAFF. (c) Model with GCAM in stage II and stage III. (d) Model with FAFF and GCAM in stage II. (e) Model with FAFF and GCAM in stage III. (f) Full model. }
  \label{Figure:ab_module}
\end{figure}

\textbf{Effectiveness of Network Modules.}
In order to verify the effectiveness of modules introduced in the proposed SUCode, we further conduct an ablation study on the components of the network. In the experiment, the baseline model is designed to replace GCAM with the basic ResBlock and remove frequency-domain related processing from FAFF. Subsequently, we introduced the frequency-domain operations into the fusion module (Table \ref{Table:netmodules}(b)), the PSNR increases from 23.000 to 23.662. This demonstrates that FAFF can model the feature interaction in the frequency domain while effectively recover high-frequency details that are severely attenuated by underwater scattering. As shown in Fig. \ref{Figure:ab_module}(b), FAFF sharpens coral textures and edge structures without amplifying noise.  In contrast, inserting GCAM alone leads to more pronounced improvements on perceptual quality, while the PSNR and SSIM gains are modest. As is shown in Fig. \ref{Figure:ab_module}(c), GCAM significantly improves color fidelity in both foreground and background regions. GCAM adaptively reweights feature channels with attention mechanism, which enhances color contrast and restores the attenuated red components. When combining FAFF and GCAM in both Stage II and Stage III, the two modules complement each other. Specifically, FAFF recovers contrast and textures while GCAM corrects color balance. This yields the highest overall performance with 23.857 PSNR, 0.925 SSIM, and 0.124 LPIPS, shown in Table \ref{Table:netmodules}(f). The visualization results in Fig. \ref{Figure:ab_module}(f) show that both sharp edges and natural colors have been preserved. This underscores the necessity of their joint integration within SUCode to effectively enhance both global color and local details.

\begin{figure}[h]
  \centering
  \includegraphics[width=0.8\linewidth]{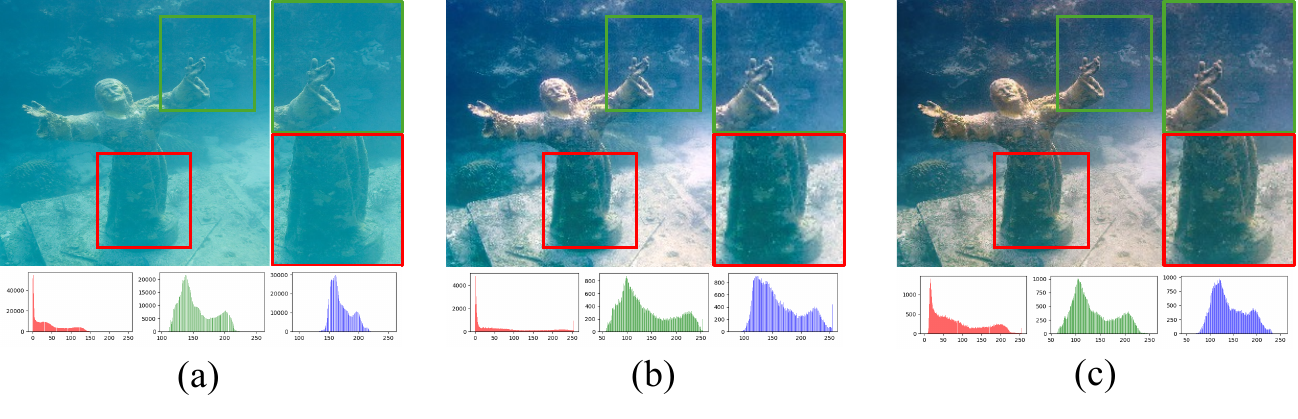}
  \caption{Visual comparison and histogram of ablation studies on the source of codebook quantization images. (a) Raw underwater images. (b) Output of codebook quantization using reference images. (c) Output of codebook quantization using raw images.}
  \label{Figure:ab_codebookimg}
\end{figure}

\begin{table}[]\centering
\caption{Ablation study of the effectiveness of network modules. The best results are highlighted in \textbf{bold}}\label{Table:netmodules}
\setlength{\tabcolsep}{4.5pt}
\scalebox{0.85}{
\begin{tabular}{c|ccc|ccc}
\toprule
     &FAFF   &GCAM-S2&GCAM-S3& PSNR$\uparrow$ & SSIM$\uparrow$ & LPIPS$\downarrow$ \\ \midrule
(a) &       &       &       &23.000&0.917 & 0.139 \\
(b) &$\surd$&       &       &23.662&0.918 & 0.139 \\
(c) &       &$\surd$&$\surd$&23.279&0.906 & 0.135 \\
(d) &$\surd$&$\surd$&       &23.009&0.911 & 0.140 \\
(e) &$\surd$&       &$\surd$&23.375&0.918 & 0.134 \\
\midrule
(f) & $\surd$ & $\surd$ & $\surd$ & \textbf{23.857} & \textbf{0.925} & \textbf{0.124}\\ 
\midrule
\end{tabular}}
\end{table}

\begin{table}[]\centering
\caption{Ablation study of the source of quantize  images. The best results are highlighted in \textbf{bold}}\label{Table:codebookimg}
\scalebox{0.85}{
\begin{tabular}{cc|ccc}
\toprule
Raw (Ours)&GT     & PSNR$\uparrow$ & SSIM$\uparrow$ & LPIPS$\downarrow$ \\ 
\midrule
$\surd$ &       & \textbf{23.857}&\textbf{0.925} & \textbf{0.124}\\
        &$\surd$& 23.216&0.913 & 0.152\\ 
\bottomrule
\end{tabular}}
\end{table}
\textbf{Effectiveness of Quantize Raw Images.}
To discuss the effectiveness of the proposed SUCode in selecting the original image during codebook construction to avoid interference from pseudo-true features, we conduct ablation experiments. For comparison, we evaluated codebook quantization using clean ground truth images in Stage I, followed by degrad-to-clean mapping in Stage II, in line with conventional VQGAN-based approaches \cite{liuLearningImageAdaptiveCodebooks2023}. The quantitative results are summarized in Table \ref{Figure:ab_codebookimg} and visual comparisons are given in Fig. \ref{Figure:ab_codebookimg}. It is observed that this setup resulted in a performance drop, with PSNR decreasing by 0.641 and SSIM by 0.012, compared to the proposed SUCode. These results highlight the limitations posed by the pseudo-GT issue in UIE, which impedes the establishment of an effective codebook. In contrast, SUCode demonstrates superior performance by building the codebook from raw underwater inputs and leveraging domain transformation in Stage III. Furthermore, the histogram of the image enhanced by SUCode is more uniform across all channels, with the red channel shifting towards medium to high intensity. This indicates that encoding the original image can better improve the network's understanding of the image as a whole, which helps to better recover color cast and expand dynamic range.

\begin{table}[]
\centering
\caption{Ablation study of codebook size. The best results are highlighted in \textbf{bold}}
\label{Table:codebooksize}
\scalebox{0.85}{
\begin{tabular}{c|ccc}
\toprule
Codebook Size   & PSNR $\uparrow$ & SSIM $\uparrow$  & LPIPS $\downarrow$ \\ 
\midrule
$128\times 128$         & 22.237 & 0.920 & 0.126\\ 
$128\times 256$         & 22.697 & 0.924 & 0.126\\ 
$256\times 256$ (SUCode) & \textbf{23.857}&\textbf{0.925} & \textbf{0.124}\\
$512\times 256$         & 23.189 & 0.914 & 0.127\\ 
$512\times 512$         & 23.065 & 0.912 & 0.141\\ 
$1024\times 512$        & 23.047 & 0.917 & 0.125\\ 
\bottomrule
\end{tabular}}
\end{table}
\textbf{Effectiveness of Codebook Size.}
We further conducted ablation experiments to verify the effect of hyper parameter settings, particularly the codebook size in the proposed SUCode. As shown in Table \ref{Table:codebooksize}, the codebook size in SUCode is set to $256\times256$ . Increasing the depth and size of the codebook beyond this setting does not improve the network's enhancement performance. For underwater images, underwater degradation patterns such as color cast, uneven illumination, and low contrast can be considered as a concentrated degradation structure. Therefore, increasing the codebook size may cause the model to have difficulty effectively utilizing all codebook vectors and introduce noisy patterns, resulting in the model's inability to learn effective quantization representations in high-dimensional space, further reducing LPIPS and SSIM. Conversely, a smaller codebook may prevent the model from fully encoding scene-related degradation features, leading to image blurring and under-reconstruction, thus reducing PSNR and SSIM metrics.

\begin{table}[]\centering
\caption{Ablation study of the accuracy of semantic masks. The semantic mask is randomly eroded and expanded within a certain pixel range. A pixel range of 0 represents using the ground truth mask.}\label{Table:pixel}
\scalebox{0.85}{
\begin{tabular}{c|ccc}
\toprule
Eroded/Expanded Range     & PSNR$\uparrow$ & SSIM$\uparrow$ & LPIPS$\downarrow$ \\ 
\midrule
0 & \textbf{23.857}&\textbf{0.925} & 0.124 \\
1-5 Pixels & 22.864 & 0.925 & \textbf{0.121} \\ 
6-10 Pixels & 23.216 & 0.921 & 0.125 \\
\bottomrule
\end{tabular}}
\end{table}

\textbf{Effectiveness of Semantic Masks.}
In the first stage of the network, SUCode construct semantic category-specific codebooks, enabling more robust image enhancement. To evaluate the effect of mask accuracy on codebook construction, we conducted a series of ablation experiments. Since the semantic masks are derived from ground-truth annotations, the primary source of noise is boundary-level annotation uncertainty rather than severe semantic errors such as class swaps or large missing regions.  Specifically, we randomly eroded or dilated the foreground region of the ground-truth mask by pixels and used the modified masks to train both the codebook quantizer and the subsequent enhancement network. The results, summarized in Table \ref{Table:pixel}, indicate that inaccuracies in the masks slightly hinder the network’s ability to model real underwater scenes, causing a minor reduction in enhancement performance. However, because the foreground and background regions of underwater images exhibit similar degradation patterns in terms of illumination and color cast, the performance drop remains within a tolerable range. 

Moreover, SUCode also incorporates several architectural designs that reduce sensitivity to imperfect semantic cues. Instead of relying on a single class-specific codebook, several quantized representations are synthesized via a learnable weight predictor, which softens hard boundary dependence and mitigates the impact of mask inaccuracies. Besides, the self-reconstruction on raw images stabilizes the discrete representation prior to enhancement, reducing the risk of error amplification from noisy semantic guidance. This confirms the robustness of the proposed SUCode to variations in input semantic information.

\textbf{Effectiveness of the Number of Codebooks.}
To analyze how the number of semantic codebooks influences enhancement quality and computational cost, an ablation is performed by merging semantically similar categories. In SUCode, multiple category-specific codebooks are used to produce quantized representations, which are then combined into a single latent representation through a weight map at each spatial location. The aggregated latent is subsequently decoded by a same decoder to obtain the enhanced output. To evaluate the impact of codebook number, we further perform category merging to 6 and 4 codebooks from the original 8, where semantically similar categories are grouped together. The results in Table \ref{Table:codenum} show that using fewer codebooks leads to a moderate degradation in enhance performance, the PSNR change from 23.857 to 22.984 to 23.216, reflecting reduced semantic specificity in the discrete representation. Meanwhile, the computational cost remains almost unchanged across different settings since the main computation is dominated by the shared encoder-decoder pathway and the fusion over codebooks introduces only a lightweight overhead. Overall, this ablation indicates that increasing the number of codebooks primarily benefits enhancement quality, while computational cost is largely stable under our aggregation-based design.

\begin{table}[]\centering
\caption{Ablation study of the number of codebooks. Of the six codebooks, we merged invertebrates and vertebrates, and then merged divers and underwater robots. Within the four codebooks, we further merged aquatic plants, ruins, and the rocks.}\label{Table:codenum}
\scalebox{0.85}{
\begin{tabular}{c|ccc|cc}
\toprule
Codebook Number   & PSNR$\uparrow$ & SSIM$\uparrow$ & LPIPS$\downarrow$ & \#Params (M) & FLOPs (G) \\ 
\midrule
8 (SUCode) & \textbf{23.857}&\textbf{0.925} & \textbf{0.124} & 61.18 & 133.21\\
6 & 22.984 & 0.917 & 0.134 & 59.67 & 131.12\\ 
4 & 23.216 & 0.921 & 0.125 & 58.23 & 129.77\\
\bottomrule
\end{tabular}}
\end{table}

\begin{figure*}[h]
  \centering
  \includegraphics[width=0.8\linewidth]{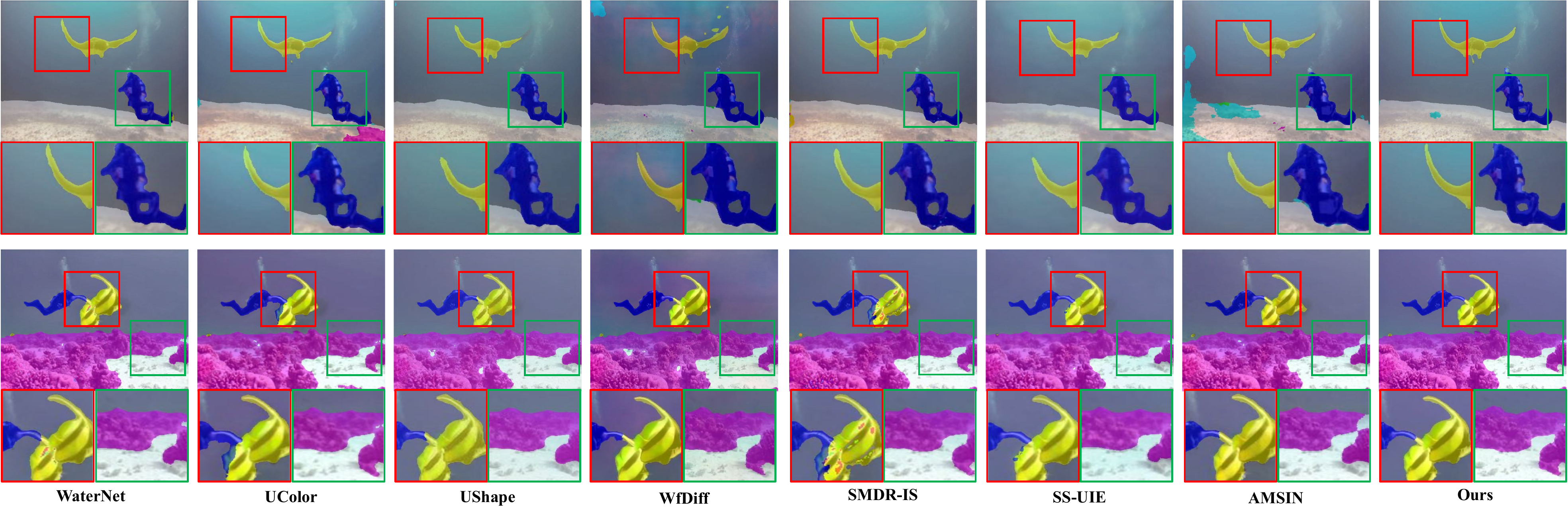}
  \caption{Visual comparison of segmentation results sampled from SUIM dataset.}
  \label{Figure:seg}
\end{figure*}

\begin{table*}[]\centering
\caption{Semantic Segmentation results on the SUIM dataset. The best results are highlighted in \textbf{bold} and the second best results are \underline{underlined}.}\label{Table:seg}
\scalebox{0.85}{
\begin{tabular}{c|ccccccc|c}
\toprule
 Method  & WaterNet\cite{liUnderwaterImageEnhancement2020} & UColor\cite{liUnderwaterImageEnhancement2021} & UShape\cite{pengUShapeTransformerUnderwater2023} & WfDiff\cite{zhaoWaveletbasedFourierInformation2024} & SMDR-IS\cite{zhangSynergisticMultiscaleDetail2024} & AMSIN\cite{quanEnhancingUnderwaterImages2024a} & SS-UIE\cite{pengAdaptiveDualdomainLearning2025} & SUCode(Ours) \\ 
\midrule
Dice Score & 62.811 & 63.434 & 60.101 & \underline{64.723} & 60.517 & 60.864 & 62.351 & \textbf{65.164} \\
mIoU & 55.888 & 54.396 & 50.791 & \textbf{66.555} & 55.677 & 52.032 & 54.443 & \underline{63.912} \\ 
\bottomrule 
\end{tabular}}
\end{table*}

\subsection{Real-World Application}
To validate the performance of SUCode in real-world applications, we further evaluate the performance of semantic segmentation on the images obtained by various UIE methods on the SUIM dataset. These results are obtained using a UNet model \cite{qiSGUIENetSemanticAttention2022} trained on the original SUIM dataset. The results are shown in Table \ref{Table:seg}. As can be seen that the proposed SUCode retains more semantic information in the image, achieving the highest Dice Score and the second best mIoU values. The segmentation results are visualized in Fig. \ref{Figure:seg} , showing that the segmentation results of the SUCode image have fewer mis-segments and are more consistent with the true outline of the object. For details in the image, such as the tips of fish fins and the edges of coral reefs, the proposed SUCode can guide the task network to achieve more refined segmentation. This can be attributed to the semantically discriminative codebook learning process, which implicitly incorporates more high-level features of underwater images into the enhanced image, facilitating downstream tasks in processing objects in the image.

\begin{figure}[h]
  \centering
  \includegraphics[width=\linewidth]{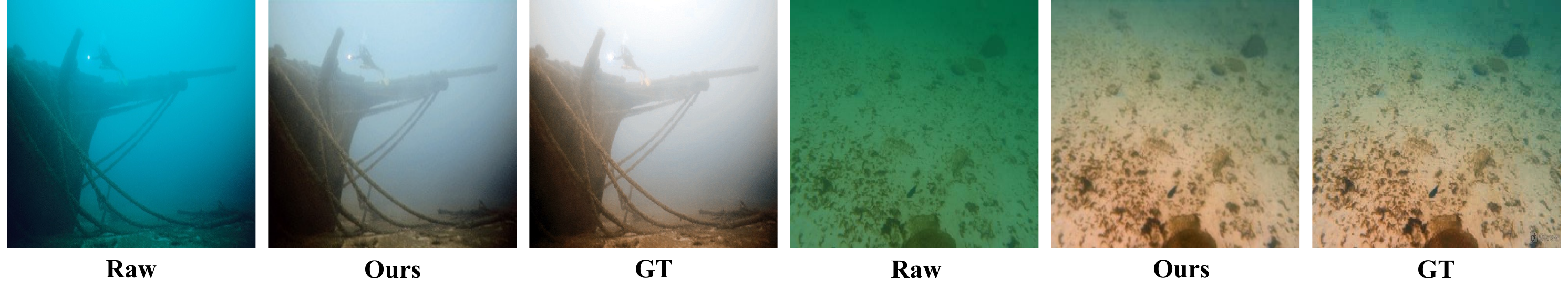}
  \caption{Visual comparison of suboptimal enhance results of the proposed SUCode. }
  \label{Figure:failurecase}
\end{figure}

\subsection{Limitations and Challenges}
Furthermore, we selected some sub-optimal enhance results from the SUIM and UIEB datasets. The original image, ground truth, and enhance results using SUCode are shown in the Fig. \ref{Figure:failurecase}. It can be seen that in cases of severe color fading or extremely low contrast, the performance of SUCode can degrade. This may be because the high-frequency details lost in the image itself cannot be supplemented by the decoder, resulting in over-smoothing and limited texture recovery, since SUCode partially relies on the structure extracted from the raw image. When there is a large difference between the brightness and darkness of an image, SUCode may not be able to enhance the shadows while preserving the details of the highlights, resulting in the loss of some information. However, in terms of visual effects, the enhanced results are close to or even exceed the ground truth. To tackle these challenges, future development include introducing robust preprocessing techniques to normalize extreme lighting conditions before enhancement, or introducing self-supervised learning methods to improve performance in highly variable underwater environments. Improvements to category-specific codebook construction mechanisms will also be considered, enabling the network to acquire more robust underwater scene codebook representations from surrounding pixels based on soft-logic semantic probability mask. Another limitation is that the input mask robustness analysis focuses on boundary perturbations. Evaluating and improving SUCode under severe semantic segmentation failures, such as category swaps and large missing regions from predicted masks remains future work.

\section{Conclusion}\label{Section:5}
We propose SUCode that handles spatially varying degradation via semantic-aware discrete codebook representations. To address inconsistencies across semantic regions, we introduce a semantic-guided pixel-level quantization strategy, enabling region-specific image feature representation. Given the ill-posed nature of UIE, we formulate enhancement as a domain conversion task in the discrete codebook space. We further design FAFF for robust feature transformation and GCAM to refine feature propagation. Experiments show that SUCode outperforms SOTA methods in image quality.

\bibliographystyle{IEEEtran}
\bibliography{Article_CodebookUIE}

\end{document}